\pdfoutput=1

\documentclass[11pt]{article}

\usepackage[]{EMNLP2022}

\usepackage{times}
\usepackage{latexsym}

\usepackage[T1]{fontenc}

\usepackage[utf8]{inputenc}

\usepackage{microtype}

\usepackage{inconsolata}

\usepackage{footnote}
\usepackage{hyperref}
\usepackage{graphicx}
\usepackage{amsmath, amssymb}
\usepackage{soul}
\usepackage{colortbl}
\usepackage{caption}
\usepackage{float}
\usepackage{booktabs}
\usepackage{hyperref}
\usepackage{multirow}
\usepackage{enumitem}
\definecolor{Gray}{gray}{0.9}

\title{ERNIE 3.0 Tiny: Frustratingly Simple Method to Improve Task-Agnostic Distillation Generalization}

\author{ \textbf{Weixin Liu, Xuyi Chen, Jiaxiang Liu, Shikun Feng, Yu Sun, Hao Tian, Hua Wu} \\ 
    Baidu Inc. \\
	\{liuweixin, chenxuyi, liujiaxiang,
	fengshikun01, sunyu02, \\ tianhao, wu\_hua\}@baidu.com \\
}

\begin{document}
\maketitle
\begin{abstract}
Task-agnostic knowledge distillation attempts to address the problem of deploying large pretrained language model in resource-constrained scenarios by compressing a large pretrained model called teacher into a smaller one called student such that the student can be directly finetuned on downstream tasks and retains comparable performance. However, we empirically find that there is a generalization gap between the student and the teacher in existing methods. In this work, we show that we can leverage multi-task learning in task-agnostic distillation to advance the generalization of the resulted student. In particular, we propose Multi-task Infused Task-agnostic Knowledge Distillation (MITKD). We first enhance the teacher by multi-task training it on multiple downstream tasks and then perform distillation to produce the student. Experimental results demonstrate that our method yields a student with much better generalization, significantly outperforms existing baselines, and establishes a new state-of-the-art result on in-domain, out-domain, and low-resource datasets in the setting of task-agnostic distillation. Moreover, our method even exceeds an 8x larger BERT$_{\text{Base}}$ on SQuAD and four GLUE tasks. In addition, by combining ERNIE 3.0, our method achieves state-of-the-art results on 10 Chinese datasets.
\end{abstract}

\section{Introduction}

Pretrained language models (PLMs) \cite{bert, roberta, electra, deberta} have achieved great success in a wide range of natural language processing tasks, however, their enormous parameters often bring challenges to serving them in real-life applications where computational resources are limited.

Knowledge distillation (KD) \cite{kd} has been widely utilized to tackle this problem. KD aims to compress a large PLM called teacher into a smaller one called student by transferring knowledge from teacher to student. In the context of PLM compression, KD is usually applied in two different settings: task-specific \cite{pkd, tang, tinybert, ernie-tiny} and task-agnostic \cite{distilbert, minilm}. The former transfers task-specific knowledge from teacher to student for a given task and often yields student with better performance than the latter, but poses one disadvantage: task-specific KD needs to be performed for every downstream task. Task-agnostic KD, on the other hand, eliminates the need of distillation for every single task by transferring general knowledge to the student in a once-for-all fashion that the student needs to be distilled only once and can be directly finetuned on downstream tasks as similar to the pretrain-finetune paradigm. A natural research question is whether we can combine the advantage of favorable downstream performance and easy-to-deploy from these two types of distillation.


Previous attempt \cite{xtremedistiltrans} injects downstream task knowledge into task-agnostic distillation by performing task-agnostic distillation on a teacher finetuned in a downstream task. Although this approach can improve the performance of the student, knowledge from a single task may not be sufficient to yield a generalizable student. In this work, we show that the downstream generalization of the student can be further improved by fusing multi-task learning (MTL) into task-agnostic distillation.
\begin{figure}[t]
      \centering
      \includegraphics[width=\linewidth]{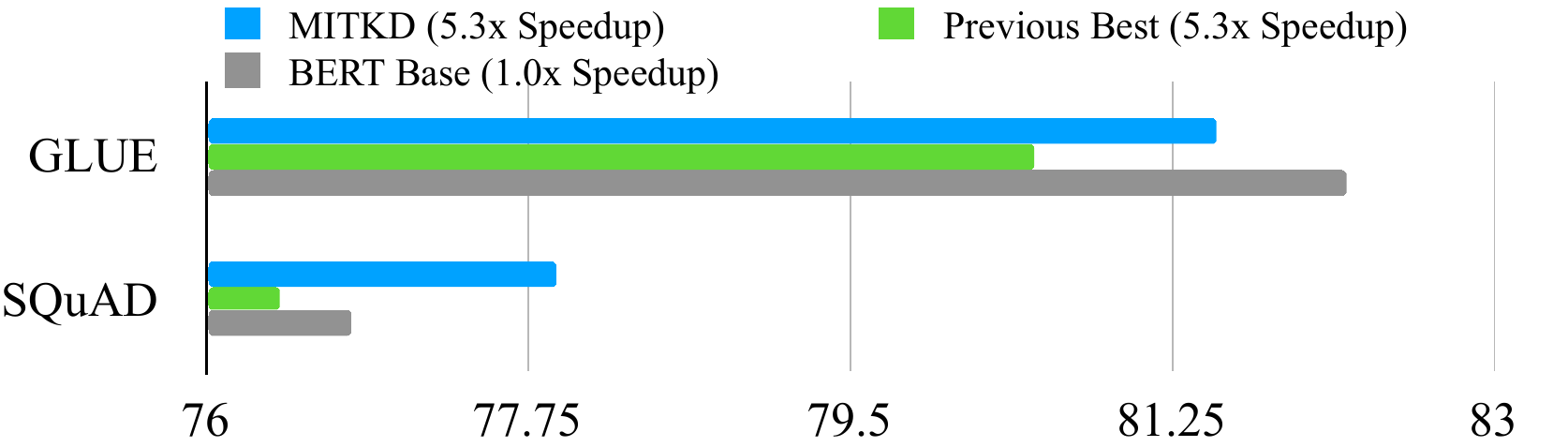}
      \vspace{-5pt}
      \caption{Performance on GLUE and SQuAD.}   
      \label{fig:muscle}
      \vspace{-20pt}
\end{figure}

Existing works in MTL \cite{mtdnn, muppet} point out that the model learns a representation that generalizes better on new tasks and domains through finetuned on multiple tasks. We propose a distillation method Multi-task Infused Task-agnostic Knowledge Distillation (MITKD) to show that the generalizable representation brought by MTL can also benefit task-agnostic distillation. In particular, we first finetune the teacher on multiple downstream tasks under the setting of MTL to learn a generalizable representation and then perform task-agnostic distillation on it. Specifically, our contribution includes:
\begin{enumerate}[noitemsep,topsep=2pt]
    \item We present a novel and simple method to combine the advantage of task-agnostic and task-specific distillation by fusing MTL into task-agnostic distillation to improve the generalization of the student.
    \item We conduct extensive experiments to verify the effectiveness of MITKD on in-domain datasets, out-domain datasets, and low-resource datasets. Empirical results show that MITKD consistently outperforms state-of-the-art baselines in all three foregoing scenarios, even outperforming an 8x larger BERT$_{\text{Base}}$ on SQuAD and four GLUE tasks shown in Figure~\ref{fig:muscle} and Table~\ref{table:glue}. Moreover, by applying MITKD on ERNIE 3.0, we obtain ERNIE 3.0 Tiny which achieves state-of-the-art results on 10 Chinese datasets.
    \item Our empirical results bring out a message that we should also pay attention on the knowledge embedded in the teacher apart from pursuing a stronger teacher, in order to improve student's downstream performance.
    
\end{enumerate}

\vspace{-3pt}
\section{Related Work}
\vspace{-0.5em}
\paragraph{Knowledge Distillation} Knowledge distillation \cite{kd} mimics the output representation of the teacher and the student to guide the student's training. In the context of PLM, task-specific methods aim to compress task-specific teacher over task-specific data \cite{pkd, tang}. \cite{mkd} strengthens the teacher through multi-task learning. \cite{BAM} compresses multiple task-specific models into one student model. 

On the other hand, task-agnostic methods aim to compress pretrained teacher in a way such that the resulted student can be easily adapted to downstream tasks via finetuning \cite{distilbert}. In addition, task-agnostic method typically performs distillation on pretraining data (i.e. the unsupervised corpora on which the teacher is pretrained). \cite{minilm, minilmv2} proposes to mimic the self-attention of student and teacher. \cite{mergedistill} compresses multiple teachers trained in different languages into a single student. XtremeDistilTrans \cite{xtremedistiltrans} distills a single-task finetuned teacher on augmented transfer data generated from unsupervised data.

\vspace{-5pt}
\paragraph{Multi-Task Learning} Multi-task learning learns multiple tasks jointly so that the knowledge learned from one task can benefit the others \cite{mtl, mtlseq}. \cite{mtdnn} shows that adding MTL at finetuning stage can boost the performance of PLM. \cite{muppet} proposes a MTL stage named pre-finetuning stage before the traditional finetuning stage. \cite{flan} combines MTL and prompting to improve zero-shot performance.

\vspace{-3pt}
\begin{savenotes}
\begin{table*}[ht]
\centering
\scalebox{0.595}{
\begin{tabular}{l|l|cccccccccccc}
\hline
\textbf{Method} & Teacher &
 CHEMPROT & ANLI & ACL-ARC & SCIERC & PARTISAN & HELPFUL & SCOTUS & LEDGAR & FinBank & Avg.\\
\hline
Domain & - & \multicolumn{1}{|c|}{Biology} & \multicolumn{1}{c|}{General}  & \multicolumn{2}{c|}{Computer Science} & \multicolumn{1}{c|}{News} & \multicolumn{1}{c|}{Review} & \multicolumn{2}{c|}{Law}  & \multicolumn{1}{c|}{Finance} & -\\
\hline
\hline
RoBERTa$_{\text{Large}}$ & -  & \textbf{86.9} & \textbf{57.8} & \textbf{82.5} & \textbf{90.6} & \textbf{87.5} & 87.7 & \textbf{77.6} & \textbf{88.6} & \textbf{90.1} & \textbf{83.3} \\
RoBERTa$_{\text{Base}}$ & - & 83.9 & 53.4 & 81.2 & 89.9 & 84.2 & 87.7 & 74.6 & 87.4 & 88.8 & 81.2\\
RoBERTa$_{\text{Base}}$ w/ MTL & - & 84.0 & 54.9 & 81.6 & 90.4 & 85.8 & \textbf{87.8} & 74.6 & 87.9 & 89.0 & 81.8\\
\hline
MiniLMv2-L & RoBERTa$_{\text{Large}}$  & 74.5 & 48.7 & 70.0 & 80.3 & 77.0 & 87.3 & 66.6 & 86.6 & 85.1 & 75.1\\
MiniLMv2-B & RoBERTa$_{\text{Base}}$ & 72.2 & 48.0 & 68.8 & 76.7 & 74.7 & 87.1 & 66.2 & 86.5 & 83.7 & 73.8 \\
XtremeDistilTrans & ELECTRA$_{\text{Base}}$* & 73.8 & \textbf{50.5} & 72.8 & 81.4 & 77.7 & 87.1 & 66.4 & 86.5 & 85.7 & 75.8\\
\rowcolor{Gray} MITKD & RoBERTa$_{\text{Base}}$ w/MTL  & \textbf{79.0} & 50.0 & \textbf{74.4} & \textbf{82.4} & \textbf{82.0} & \textbf{87.5} & \textbf{71.2} & \textbf{86.8} & \textbf{85.9} & \textbf{77.7}\\
\hline
\end{tabular}
}
\caption{Out-domain results on the development sets. All results are produced by us using their publically available checkpoints. For ANLI, we merge the train sets and dev sets at all three rounds into one train set and dev set, and use them for training and evaluation. * means that ELECTRA$_{\text{Base}}$ is finetuned on MNLI.}\label{table:out_domain}
\vspace{-1.5em}
\end{table*}
\end{savenotes}

\section{Method}
\vspace{-7pt}
In this section, we first categorize the existing works in task-agnostic distillation into two different types, then we propose MITKD and show the difference between it and these existing works. 
\vspace{-3pt}
\paragraph{Vanilla Task-agnostic Distillation} Vanilla task-agnostic distillation transfers general knowledge from pretrained teacher to student without leveraging any downstream knowledge. 

\vspace{-3pt}
\paragraph{Single-task Enhanced Task-agnostic Distillation} Single-task enhanced task-agnostic distillation exploits downstream knowledge by finetuning the teacher on a single task and performing distillation on it. For example, XtremeDistilTrans \cite{xtremedistiltrans} develops a cumbersome distillation method that first intensively searches for a source task inducing better transferability, then finetunes the pretrained teacher in the source task. After that, it performs task-agnostic distillation on the finetuned teacher with augmented transfer data generated from unsupervised data and a previously task-agnostic distilled student as a warm start.


\vspace{-3pt}
\paragraph{Multi-task Infused Task-agnostic Distillation (MITKD)} While previous work leverages task knowledge through carefully hunting for a task inducing better transferability, we argue that we can simply utilize the power of MTL to infuse task knowledge into task-agnostic distillation to improve student's generalizability. In particular, we propose a two-stage distillation method: we first finetune a pretrained teacher on multiple tasks and then perform vanilla task-agnostic distillation on it. Unlike previous work which requires searching for the best transferable task and distilling on augmented transfer data, our method simply applies MTL to the teacher and only needs to distill on pretraining data as what vanilla task-agnostic distillation does. Moreover, our empirical results illustrate that MITKD does not only inject task knowledge to the distillation, but more importantly, brings in the generalization to improve the downstream performance of the student dramatically.

\vspace{-3pt}
\section{Experiment}
\vspace{-7pt}
\paragraph{Experiment Setup}
First, we finetune the teacher with MTL. In particular, we adopt the base version of Muppet's \cite{muppet} released checkpoint \footnote{\url{https://huggingface.co/facebook/muppet-roberta-base}} as our finetuned teacher. It is obtained by finetuning a pretrained RoBERTa$_{\text{Base}}$ model \cite{roberta} on around 50 tasks jointly. The datasets used for the the MTL finetuning are listed in Appendix~\ref{app:muppet_dataset}. 

Then, we perform task-agnostic distillation on the multi-task finetuned teacher using a classic task-agnostic distillation method MiniLMv2~\cite{minilmv2} which mimics a more fine-grained version of self-attention between the teacher and the student. Following MiniLMv2, we utilizes the pretraining datasets used in RoBERTa for the distillation data. All the students in the experiment are 6-layer transformers with hidden size of 384, intermediate size of 1536, and 12 attention heads. All the results in the experiment section are reported on the development set and are an average of 4 runs. We also listed all the hyperparameters used in the experiment in Appendix~\ref{app:hyperparam}. 

\vspace{-4pt}
\paragraph{Baselines}
We select the classic task-agnostic distillation MiniLMv2~\cite{minilmv2} as our baseline for \textit{vanilla task-agnostic distillation}. It utilizes the same task-agnostic distillation algorithm as us but is distilled from a larger teacher RoBERTa$_{\text{Large}}$. In order to ablate the effectiveness of our method, we also reproduce a MiniLMv2 with RoBERTa$_{\text{Base}}$ which is essentially the same teacher as ours but without MTL training. We denote the one distilled from RoBERTa$_{\text{Large}}$ as MiniLMv2-L and the other as MiniLMv2-B. As for \textit{single-task enhanced task-agnostic distillation}, we select the state-of-the-art method XtremeDistilTrans \cite{xtremedistiltrans} as the baseline. In particular, it utilizes a ELECTRA$_{\text{Base}}$ \cite{electra} finetuned on MNLI~\cite{mnli} as the teacher and performs distillation from it on a student previously task-agnostic distilled.

\vspace{-4pt}
\paragraph{Tasks and Datasets}
To demonstrate how MTL can benefit task-agnostic distillation, we evaluate the student on those tasks utilized in the MTL stage of the teacher and those tasks which are not used in MTL. For simplicity, we call the former in-domain tasks and the latter out-domain tasks. In order to verify the generalization improvement brought by MITKD on out-domain datasets, we experiment with nine datasets from seven domains. Refer to Appendix~\ref{app:out_domain} for the dataset details. As for the evaluation on in-domain tasks, we select GLUE benchmark \cite{glue} and SQuAD 2.0 \cite{squad2} among the 50 tasks trained in the MTL stage of the teacher.
\begin{savenotes}
\begin{table*}[!th]
    \centering
    \scalebox{0.7}{
    \begin{tabular}{l|l|cccccccccc}
        \hline
        \textbf{Method} & Teacher &
         MNLI & MRPC & QNLI & QQP & RTE & CoLA & SST & SQuADv2	& Avg.\\
        \hline
        BERT$_{\text{Base}}$ & - & 84.5 & 87.3 & 91.7 & 91.3 & 68.6 & 58.9 & 93.2 & 76.8 & 81.5\\
        RoBERTa$_{\text{Large}}$ & - & \textbf{90.2} & 90.9 & \textbf{94.7} & \textbf{92.2} & 86.6 & \textbf{68.0} & 96.4 & \textbf{89.4} & \textbf{88.6} \\
        RoBERTa$_{\text{Base}}$ & - & 87.6 & 90.2 & 92.8 & 91.9 & 78.7 & 63.6 & 94.8 & 83.7 & 85.4\\
        RoBERTa$_{\text{Base}}$ w/ MTL & - & 88.1 & \textbf{91.7} & 93.3 & 91.9 & \textbf{87.8} & 64.6$^{\dagger}$  & \textbf{96.7} & 84.7$^{\dagger}$ & 87.4\\
        \hline
        MiniLMv2-L & RoBERTa$_{\text{Large}}$ & 84.4 & 88.7 & 90.9 & 90.8 & 69.9 & \textbf{42.6} & 92.0 & 76.4 & 79.5\\
        MiniLMv2-B & RoBERTa$_{\text{Base}}$ & 83.4 & 86.4 & 90.2 & 90.7 & 59.6 & 36.6 & 91.9 & 75.5 & 76.8\\
        XtremeDistilTrans & ELECTRA$_{\text{Base}}$* & 84.5 & 89.0 & 90.2 & 90.4 & 77.3 & 40.6$^{\dagger}$ & 91.6 & 74.4 & 79.8\\
        \rowcolor{Gray} MITKD & RoBERTa$_{\text{Base}}$ w/ MTL & \textbf{85.1} & \textbf{89.6} & \textbf{91.7} & \textbf{91.0} & \textbf{77.6} & 41.8 & \textbf{93.6} & \textbf{77.9} & \textbf{81.0}\\
        \hline
    \end{tabular}
    }
    \captionof{table}{In-Domain results on the development sets of GLUE and SQuAD 2.0 . We quote the result for RoBERTa$_{\text{Base}}$ w/ MTL, MiniLMv2-L and XtremeDistillTrans from their papers. The result for BERT is taken from \cite{minilm}. The results for RoBERTa are taken from their github\footnote{\url{https://github.com/facebookresearch/fairseq/tree/main/examples/roberta}}. The number with $\dagger$ indicates that the number is missing and we report our reproduced number using the publicly available checkpoint. We report F1 for SQuAD.\label{table:glue}}
    \vspace{-1.5em}
\end{table*}
\end{savenotes}

\vspace{-4pt}
\paragraph{Out-domain Generalization} As MTL brings better generalization for new tasks and domains to the teacher \cite{muppet}, we empirically show that it also boosts the generalization of the student through distillation by evaluating the downstream performance of the student on out-domain tasks. In particular, we compare our method with the mentioned baselines in Table~\ref{table:out_domain}. Experimental results illustrate that our method significantly outperforms other baselines, demonstrating the generalization brought by our method. Comparing XtremeDistilTrans and MiniLMv2-B, we can see that introducing task knowledge to distillation can improve the performance of the student. Moreover, MITKD further brings in multi-task knowlegde and generalization led by MTL, pushing the improvement by 2.0 average points. Together, we observe that adding MTL into task-agnostic distillation results in an impressive improvement of 3.9 points on out-domain tasks, compared to the vanilla task-agnostic distillation baseline MiniLMv2-B. In addition, it even exceeds MiniLMv2-L by 2.6 points.

\vspace{-4pt}
\paragraph{In-domain Performance} Besides the generalization improvement on out-domain datasets, MITKD also improves the performance on in-domain tasks. Table~\ref{table:glue} shows the dev set result on GLUE and SQuAD 2.0. MITKD achieves state-of-the-art results on most tasks in GLUE and SQuAD 2.0, exceeding the best baseline by a margin of 1.2 points. It even outperforms an 8x larger BERT$_{\text{Base}}$ on four GLUE tasks and SQuAD 2.0 while being 5.3x faster than BERT$_{\text{Base}}$.\footnote{Refer to Appendix~\ref{app:speed} for the details of speedup and model size calculation.}

\begin{table}[h!]
\centering
\scalebox{0.6}{
\begin{tabular}{lcccc}
\hline
Method &
QNLI & CHEMPROT & LEDGAR & SCIREC	\\
\hline
Domain & In & Biology & Law & Comp. Sci. \\
\hline\hline
\rowcolor{Gray}
\# of train data (1\%) & 1047 & 41 & 600 & 32 \\
\hline
MiniLMv2-L  & 80.2 & \textbf{33.5} & 21.2 & \textbf{47.0} \\
MiniLMv2-B  & 78.6 & \textbf{33.5} & 20.5 & \textbf{47.0} \\
XtremeDsitilTran & 84.1 & \textbf{33.5} & 18.3 & \textbf{47.0}  \\
MITKD  & \textbf{85.2} & \textbf{33.5} & \textbf{36.6} & \textbf{47.0}\\ 
\hline\hline
\rowcolor{Gray}
\# of train data (10\%) & 10474 & 416 & 6000 & 321\\
\hline
MiniLMv2-L  & 86.6 & 36.9 & 74.2 & 50.0 \\
MiniLMv2-B  & 84.7 & 36.3 & 73.9 & 47.6 \\
XtremeDistilTran & 87.7 & 40.0 & 66.7 & 52.8 \\
MITKD  & \textbf{89.3} & \textbf{50.2} & \textbf{77.8} & \textbf{54.8} \\ 
\hline\hline
\rowcolor{Gray}
\# of train data (50\%) & 52371 & 2084 & 30000 & 1609\\
\hline
MiniLMv2-L  &  89.8 & 63.5 & 83.8 & 71.7 \\
MiniLMv2-B  & 88.8 & 61.7 & 83.4 & 69.8\\
XtremeDsitilTran &  90.1 & 66.3 & 83.7 & 75.0 \\
MITKD & \textbf{91.0} & \textbf{74.8} & \textbf{84.8} & \textbf{78.5}\\ 
\hline
\end{tabular}
}
\caption{\label{table:less_data_resut} Transferability over different dataset scale settings.}
\vspace{-1.5em}
\end{table}

\vspace{-4pt}
\paragraph{Performance on Low-resource Tasks} MITKD also brings improvement on low-resource tasks in both in-domain and out-domain. To demonstrate that, we select datasets from various domains and vary their training data size to 1\%, 10\%, and 50\%. In particular, we use QNLI for in-domain, CHEMPROT for biology, LEDGAR for legal, SCIREC for computer science. Note that only QNLI has been used for MTL training among these tasks. The results are presented in Table~\ref{table:less_data_resut}, from which we can see that MITKD consistently outperforms other baselines. Recall that XtremeDistilTrans is obtained by distilling from an MNLI-finetuned teacher. It performs relatively well in QNLI which is similar to MNLI, but fails to transfer well and even underperforms the vanilla task-agnostic distillation baseline MiniLMv2-B in the 1\% and 10\% settings of legal (LEDGAR). On the other hand, MITKD consistently performs the best, demonstrating that MTL brings significantly better generalization to the student on low-resource tasks than single task finetuning.

\begin{table}[h]
    \vspace{-5pt}
    \centering
    \scalebox{0.61}{
    \begin{tabular}{l|c|cc}
        \hline
        \textbf{Method} & Teacher &
        In-Domain & Out-Domain \\
        \hline
        RoBERTa$_{\text{Large}}$ & - & \textbf{88.6} & \textbf{83.3} \\
        RoBERTa$_{\text{Base}}$ & - & 85.4 & 81.2 \\
        RoBERTa$_{\text{Base}}$ w/ MTL & - & 87.4 & 81.8 \\
        \hline
        MiniLMv2-L &        \multicolumn{1}{|l|}{RoBERTa$_{\text{Large}}$}             & 79.5 & 75.1\\
        MiniLMv2-B &           \multicolumn{1}{|l|}{RoBERTa$_{\text{Base}}$} & 76.8 & 73.8\\
        \rowcolor{Gray} MITKD &           \multicolumn{1}{|l|}{RoBERTa$_{\text{Base}}$ w/ MTL }      & \textbf{81.0} & \textbf{77.7} \\
        \hline
    \end{tabular}
    }
    \captionof{table}{Summary of average scores on in-domain and out-domain tasks.  \label{table:muscle}}
    \vspace{-17pt}
\end{table}
\vspace{-4pt}
\paragraph{Larger Teacher or Better Knowledge?} There are many ways to improve the performance of the student. One straightforward way is to enlarge the teacher to obtain a stronger teacher with better downstream performance: for example, MiniLMv2-L outperforms MiniLMv2-B by simply switching to a larger teacher, shown in Table~\ref{table:muscle}. However, several works \cite{TA, dynadistill} witness failure when the size gap between student and teacher are too large. On the other hand, we propose to enhance the teacher with more generalizable knowledge through MTL. Table~\ref{table:muscle} shows that although our teacher has inferior performance compared to RoBERTa$_{\text{Large}}$, the produced student shows better performance on in-domain and out-domain tasks. This suggests that besides chasing after a larger and stronger teacher, we should also pay attention to the knowledge embedded in the teacher.

\begin{savenotes}
\begin{table*}[!ht]
\centering
\scalebox{0.6}{
\begin{tabular}{l|cccccc|cc|ccc|c}
\hline
Method &
AFQMC & TNEWS & IFLYTEK & OCNLI & CLUEWSC & CSL & CANLI & SHOPPING10 & BUSTM & EPRSTMT & CSLDCP & Avg. \\
\hline
Domain & \multicolumn{6}{c}{In-Domain} & \multicolumn{2}{|c|}{Out-Domain}  & \multicolumn{3}{c|}{Low-Resource} & All\\
\hline
TinyBERT, Chinese & 69.1 & 54.0 & 39.7 & 69.6 & \textbf{70.1} & 75.1 & 93.1 & 93.6 & 71.3 & 78.6 & 34.1 & 68.0\\
ERNIE 3.0 Tiny & \textbf{72.4} & \textbf{55.4} & \textbf{50.6} & \textbf{74.4} & 68.4 & \textbf{76.5} & \textbf{98.2} & \textbf{94.5} & \textbf{72.3} & \textbf{87.0} & \textbf{45.1} & \textbf{72.3}\\
\hline
\end{tabular}
}
\caption{\label{table:chinese_datasets} Performance on Chinese Datasets. All results listed are reported on dev set.}
\vspace{-1.5em}
\end{table*}
\end{savenotes}

\paragraph{Performance on Chinese Datasets} To verify the effectiveness of MITKD on Chinese datasets, we apply MITKD on ERNIE 3.0 \cite{ernie3p0} to produce ERNIE 3.0 Tiny. In particular, we follow ERNIE 3.0 training process to reproduce a Large version of ERNIE 3.0 and use it as the teacher. It is a 24-layer transformer with hidden size of 1024, intermediate size of 4096, and 16 attention heads. 28 datasets are selected for MTL finetuning the teacher and are listed in Appendix~\ref{app:chinese_dataset}.  All the students in this experiment are 4-layer transformers with hidden size of 312 and 12 attention heads. The finetuning hyperparameters for the students are listed in Appendix~\ref{app:chinese_hyperparam}. We compare ERNIE 3.0 Tiny with the Chinese version of TinyBERT \footnote{\url{https://github.com/huawei-noah/Pretrained-Language-Model/tree/master/TinyBERT}} \cite{tinybert} on the in-domain, out-domain and low-resource datasets and list out the result in Table~\ref{table:chinese_datasets}. Refer to Appendix~\ref{app:chinese_finetune_datasets} for the dataset details. ERNIE 3.0 Tiny outperforms the baseline by a margin of 4.3 average points, establishing a new state-of-the-art result.

\vspace{-6pt}
\section{Conclusion}
\vspace{-0.5em}
In this work, we propose a simple method to improve task-agnostic distillation generalization by leveraging MTL. The teacher is first augmented by MTL and then distilled. Empirical results show that our method outperforms several baselines on in-domain, out-domain and low-resource tasks.

\section{Limitation}
Compared with vanilla task-agnostic distillation, MITKD has an additional multi-task finetuning stage which may require additional computation resources.

\bibliography{anthology,custom}
\bibliographystyle{acl_natbib}

\appendix

\newpage
\section{Appendix}

\subsection{Out-domain Datasets}\label{app:out_domain}
We evaluate our model on CHEMPROT \cite{chemprot} for Biology, ANLI \cite{anli} for web, ACL-ARC \cite{acl-arc} and SCIERC \cite{scierc} for computer science, PARTISAN \cite{hyperpartisan} for news, HELPFUL \cite{helfulness} for review, SCOTUS\cite{scotus} and LEDGAR \cite{ledgar} for law, and FinBank \cite{finbank} for finance.

\subsection{Hyperparameters} \label{app:hyperparam}
We use Adam \cite{adam} optimizer with $\beta_{1}$ of 0.9 and $\beta_{2}$ of 0.999 for all tasks. We use linear warm up over the first 10\% of training steps and linear decay for the rest of training steps. The dropout rate is set to 0.1. The weight decay is 0.01.
\paragraph{SQuAD 2.0} The learning rate ranges from \{3e-5, 6e-5, 8e-5, 9e-5\}. The batch size is set to 32. The training epoch is set to 3. 
\paragraph{Others}\label{app:hyperparam_other} For all other tasks, the epoch ranges from \{3, 5, 10\}, the batch size ranges from \{16, 32, 48\}, learning rate ranges from \{1e-5, 2e-5, 5e-5\}. For CoLA, we finetune it with training epoch of 25.

\subsection{Model size and Speed Calculation} \label{app:speed} 
When calculating the size of a model, we count the number of all the learnable parameters in the model except the parameters of the embeddings. As for the speedup, we quote the number from \cite{minilmv2}.

\subsection{Datasets Used for Fintuning Teacher}\label{app:muppet_dataset}
\begin{enumerate}
    \item MNLI \cite{mnli}
    \item CoLA \cite{cola}
    \item SST \cite{sst2}
    \item MRPC \cite{mrpc}
    \item QQP \cite{qqp}
    \item QNLI \cite{squad}
    \item RTE \cite{rte}
    \item WNLI \cite{wsc}
    \item Bool Q \cite{boolq}
    \item MultiRC \cite{multirc}
    \item ReCoRD \cite{zhang2018record}
    \item WIC \cite{wic}
    \item WSC \cite{wsc}
    \item CB \cite{cb}
    \item COPA \cite{copa}
    \item AG News \cite{ag_news}
    \item IMDB \cite{imdb}
    \item SNLI \cite{snli}
    \item HANS \cite{hans}
    \item Rotten Tomatoes \cite{rotten_tomatoes}
    \item Yelp Polarity \cite{ag_news}
    \item Eraser Multi RC \cite{eraser_multi_rc}
    \item Wiki QA \cite{wiki_qa}
    \item Trec \cite{trec1, trec2}
    \item SciTail \cite{scitail}
    \item CNN Daily Mail \cite{cnn_daily_mail}
    \item Billsum \cite{billsum}
    \item XSUM \cite{xsum}
    \item Aeslc \cite{aeslc}
    \item Multinews \cite{multinews}
    \item Math QA \cite{math_qa}
    \item Openbook QA \cite{openbook_qa}
    \item SWAG \cite{swag}
    \item HellaSWAG \cite{hella_swag}
    \item RACE \cite{race}
    \item CommonSense QA \cite{commonsense_qa}
    \item Cosmos QA \cite{cosmos_qa}
    \item AI2 ARC - Easy \cite{ai2_arc}
    \item AI2 ARC - Challenge \cite{ai2_arc}
    \item SCIQ \cite{sciq}
    \item SQUAD \cite{squad}
    \item NQ \cite{nq}
    \item DROP \cite{drop}
    \item Hotpot \cite{hotpot}
    \item Trivia QA \cite{trivia_qa}
\end{enumerate}

\begin{savenotes}
\begin{table*}[!ht]
\centering
\scalebox{0.83}{
\begin{tabular}{l|ccccccc}
\hline
Task & AFQMC & TNEWS & IFLYTEK & OCNLI & CLUEWSC & CSL	\\
\hline
Epoch  & 3 & 3 & 3 & 5 & 50 & 5\\
Dropout Rate & 0.1 & 0.1 & \{0.0, 0.1\} & 0.1 & \{0.0, 0.1\} & 0.1\\
\hline
\end{tabular}
}
\caption{\label{table:chinese_hyparameters} Epoch and dropout rate settings for chinese datasets.}
\end{table*}
\end{savenotes}

\subsection{Datasets Used for Fintuning ERNIE 3.0 Titan}\label{app:chinese_dataset}
\begin{enumerate}
    \item XNLI \cite{xnli}
    \item OCNLI \cite{ocnli}
    \item KUAKE-QQR \cite{cblue}
    \item KUAKE-QTR \cite{cblue}
    \item CMNLI \cite{CLUE}
    \item PAWS-X \cite{paws_x}
    \item CHIP-STS \cite{cblue}
    \item CSL \cite{csl}
    \item CCKS2018-Task3 \footnote{\url{http://www.sigkg.cn/ccks2018/?page_id=16}}
    \item CHIP2019-Task2 \footnote{\url{http://www.cips-chip.org.cn:8000/evaluation}}
    \item ZHONGCE: It is an internal semantic textual similarity dataset, consisting of 1025327 training examples and 8803 evaluation examples.
    \item QBQTC \footnote{\url{https://github.com/CLUEbenchmark/QBQTC}}
    \item LCQMC \cite{lcqmc}
    \item BQ \cite{bq}
    \item AFQMC \cite{CLUE}
    \item WAIMAI \footnote{\url{https://github.com/SophonPlus/ChineseNlpCorpus/tree/master/datasets/waimai_10k}}
    \item WEIBO \footnote{\url{https://github.com/SophonPlus/ChineseNlpCorpus/tree/master/datasets/weibo_senti_100k}}
    \item NLPC2014-Task2 \footnote{\url{http://tcci.ccf.org.cn/conference/2014/pages/page04_dg.html}}
    \item SemEval2016-Task5-PHNS \cite{se16-absa}
    \item SemEval2016-Task5-CAME \cite{se16-absa}
    \item THUCNEWS \footnote{\url{http://thuctc.thunlp.org}}
    \item TNEWS \cite{CLUE}
    \item IFLYTEK \cite{CLUE}
    \item SOGOUNEWS \cite{ag_news}
    \item CHIP-CTC \cite{cblue}
    \item CNSE \cite{cnss}
    \item CNSS \cite{cnss}
    \item CLUEWSC \cite{CLUE}
\end{enumerate}

\subsection{Chinese Evaluation Datasets}\label{app:chinese_finetune_datasets}
We select the classification tasks in CLUE Benchmark 
 \cite{CLUE} for the evaluation on in-domain datasets, and CANLI \cite{canli} and SHOPPING10\footnote{\url{https://github.com/SophonPlus/ChineseNlpCorpus/tree/master/datasets/online_shopping_10_cats}} for the out-domain datasets. In addition, we select BUSTM, EPRSTMT and CSLDCP from FewCLUE \cite{fewclue}, a Chinese few-shot learning benchmark, for the low-resource datasets. For the low-resource datasets, we merge the train sets and dev sets from all sub-parts into one train set and dev set, and use them for training and evaluation.

\subsection{Hypaerparameters for Chinese Datasets}\label{app:chinese_hyperparam}
For the in-domain datasets, we follow the hyperparameters settings in PaddleNLP GitHub repository\footnote{\url{https://github.com/PaddlePaddle/PaddleNLP/tree/develop/examples/benchmark/clue}}. In particular, the batch size ranges from \{16, 32, 64\}, learning rate ranges from \{1e-5, 2e-5, 3e-5, 5e-5\}. The epoch and dropout rate for each task are listed in Figure~\ref{table:chinese_hyparameters}. All other settings such as optimizer, and learning rate are the same as those in Appendix~\ref{app:hyperparam}. For all other tasks, we use the same hyperparameters in Appendix~\ref{app:hyperparam_other}.

\end{document}